\documentclass[10pt,twocolumn,letterpaper]{article}
\usepackage[accsupp]{axessibility}
\usepackage{cvpr}              
\usepackage{multirow}
\def\vs{\emph{vs.}}

\usepackage[dvipsnames]{xcolor}

\definecolor{cvprblue}{rgb}{0.21,0.49,0.74}
\usepackage[pagebackref,breaklinks,colorlinks,citecolor=cvprblue]{hyperref}
\def\paperID{1208} 
\def\confName{CVPR}
\def\confYear{2024}

\title{CFPL-FAS: Class Free Prompt Learning for Generalizable Face Anti-spoofing}
\newcommand*\samethanks[1][\value{footnote}]{\footnotemark[#1]}

\author{
	Ajian Liu$^{\rm 1}$, 
	Shuai Xue$^{\rm 2}$, 
	Jianwen Gan$^{\rm 3}$, 
	Jun Wan$^{\rm 1, 4, 5}$\thanks{Corresponding author.}, 
	Yanyan Liang$^{\rm 4}$\samethanks\\
	Jiankang Deng$^{\rm 6}$, 
	Sergio Escalera$^{\rm 7}$, 
	Zhen Lei$^{\rm 1,5,8}$ \\
	$^{\rm 1}$MAIS, CASIA, China; 
	$^{\rm 2}$BIT, Zhuhai; 
	$^{\rm 3}$GC\&UKLIS, Wuzhou University;
	$^{\rm 4}$M.U.S.T, Macau \\
	$^{\rm 5}$SAI, UCAS, China; 
	$^{\rm 6}$ICL, UK; 
	$^{\rm 7}$CVC, Spain; 
	$^{\rm 8}$CAIR, HKISI, CAS \\
	\tt\footnotesize
	$^1$\{ajian.liu,jun.wan\}@ia.ac.cn, 
	\tt\footnotesize
	$^4$yyliang@must.edu.mo 
	\tt\footnotesize
	$^1$zlei@nlpr.ia.ac.cn 
}

\begin{document}
\maketitle
\begin{abstract}  
	Domain generalization (DG) based Face Anti-Spoofing (FAS) aims to improve the model's performance on unseen domains. Existing methods either rely on domain labels to align domain-invariant feature spaces, or disentangle generalizable features from the whole sample, which inevitably lead to the distortion of semantic feature structures and achieve limited generalization. In this work, we make use of large-scale VLMs like CLIP and leverage the textual feature to dynamically adjust the classifier's weights for exploring generalizable visual features. Specifically, we propose a novel Class Free Prompt Learning (\textbf{CFPL}) paradigm for DG FAS, which utilizes two lightweight transformers, namely Content Q-Former (CQF) and Style Q-Former (SQF), to learn the different semantic prompts conditioned on content and style features by using a set of learnable query vectors, respectively. Thus, the generalizable prompt can be learned by two improvements: (1) A Prompt-Text Matched (\textbf{PTM}) supervision is introduced to ensure CQF learns visual representation that is most informative of the content description. (2) A Diversified Style Prompt (\textbf{DSP}) technology is proposed to diversify the learning of style prompts by mixing feature statistics between instance-specific styles. Finally, the learned text features modulate visual features to generalization through the designed Prompt Modulation (\textbf{PM}). Extensive experiments show that the CFPL is effective and outperforms the state-of-the-art methods on several cross-domain datasets.
\end{abstract} 

\section{Introduction}
\label{sec:intro}
Face Anti-Spoofing (FAS) is an important step in protecting the security of face recognition systems from print-attack~\cite{Zhang2012A}, replay-attack~\cite{Chingovska_BIOSIG-2012} and mask-attack~\cite{liu2022contrastive,fang2023surveillance}. Despite the existing presentation attack detection (PAD) methods~\cite{Liu2018Learning,george2019deep,zhang2020face,liu2020disentangling,yu2020nasfas,liu2022disentangling,Liu_2023_CVPR} obtain remarkable performance in intra-dataset experiments where training and testing data are from the same domain, their performance severely degraded in cross-dataset experiments due to large distribution discrepancies among different domains.
{\tiny \begin{figure}[t]
		\centering
		\includegraphics[width=1.0\linewidth]{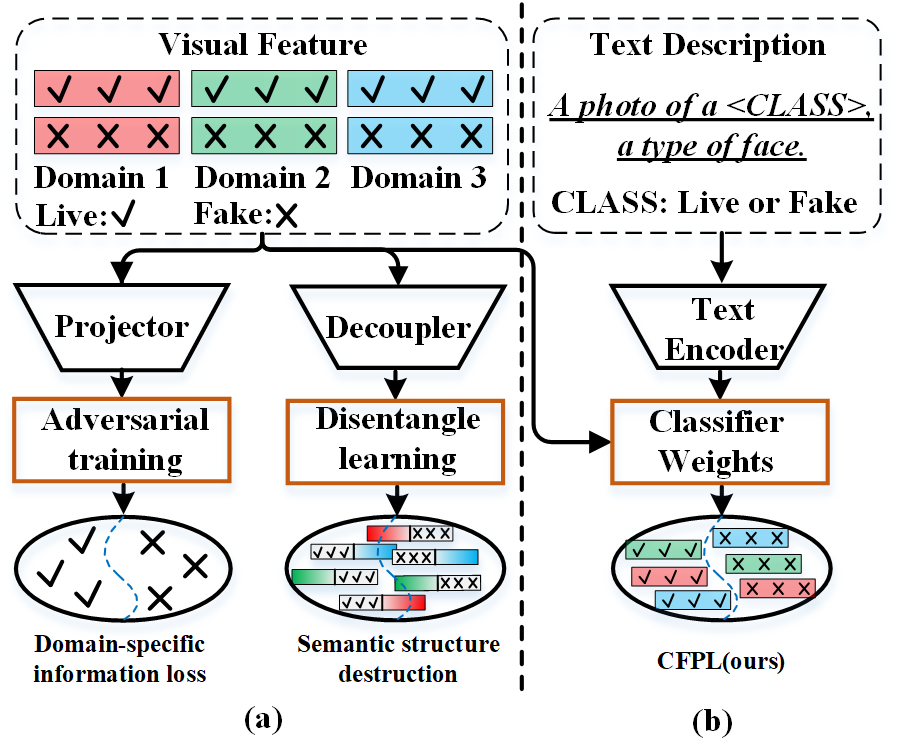}
		\caption{Comparison with existing DG FAS methods. (a) the previous methods either rely on a projector to align domain-invariant feature spaces with adversarial training, or disentangle generalizable features from the whole sample with a decoupler, which inevitably leads to the distortion of semantic structures and achieves limited generalization. (b) Our CFPL framework is built on CLIP to learn generalized visual features by using the text features as weights of the classifier.}
		\label{fig:dg_fas}
	\end{figure}
}
Domain Generalization (DG) based FAS aims to mitigate the impact of distribution discrepancy by accessing multiple domains. As shown in Fig.~\ref{fig:dg_fas} (a) (left), a typical strategy~\cite{Rui2019Multi,shao2020regularized,jia2020single,wang2022domain} is rely on domain labels to learn a domain-invariant feature space by adversarial training, which is also generalized to unseen domains. While it is difficult to seek a compact and generalized feature space for all domains. Even, there is no guarantee that such a feature space exists among multiple domains due to large distribution discrepancies. Considering the manually labeled domain labels are arbitrary and subjective, which cannot truly reflect the diversity of samples in a domain, as shown in Fig.~\ref{fig:dg_fas} (a) (right), another strategy~\cite{chen2021generalizable,liu2021adaptive,zhou2023instance} is based on the instance to disentangle or generate generalized features from liveness-irrelevant features by disentangled representation learning.
Based on the above analysis, these methods either rely on domain labels to align domain-invariant feature spaces, or disentangle generalizable features from instance-specific and liveness-irrelevant features, which inevitably leads to the distortion of semantic feature structures and achieve limited generalization. 

Rethinking the reason for the poor generalization of FAS can be attributed to the liveness-irrelevant features interfering with the classifier's recognition of spoofing clues. If the weight of the classifier can be dynamically adjusted based on sample instances, such as, to weaken the interference factors and strengthen the generalized features, it will effectively improve its generalization. 
Inspired by large vision-language models like \emph{CLIP}~\cite{radford2021learning}, which jointly trains an image encoder and a text encoder to predict the correct of pairings (image, text), As shown in Fig.~\ref{fig:dg_fas} (b), we generate corrective text features by a high-capacity text encoder, which allows open-set visual concepts and broader semantic spaces compared to discrete domain labels. 
How to learn generalizable prompts for text encoder to adaptively adjust the classifier‘s weights? Unlike general DG task, which have domain information such as `This image belongs to Cartoon/Sketch/Art Painting/Photo‘, a FAS dataset is regarded as a domain, usually named with the publishing agency, such as MSU~\cite{wen2015face}, CASIA~\cite{Zhang2012A}, Idiap~\cite{Chingovska_BIOSIG-2012}, or OULU~\cite{Boulkenafet2017OULU}, without providing any valuable domain semantic information to assist generalizable prompt learning. 

In this work, without relying on domain semantics, the generalizable prompts can be based on image content and style learned by reducing their correlation~\cite{wang2022domain,zhou2023instance}. Based on this determination, inspired by BLIP-2~\cite{li2023blip} and TGPT~\cite{tan2023compound}, 
we design two lightweight transformers, namely Content Q-Former (CQF) and Style Q-Former (SQF), to learn the expected prompts conditioned on content and style features by using a set of learnable query vectors, respectively. To further ensure CQF can learn to extract the visual representation that is most informative of the content description, we introduce a Prompt-Text Matched (\textbf{PTM}) supervision to optimize the learning of content prompts, where each sample's content description is generated by template description. Due to the inability to accurately describe style information in text, instead, we propose a Diversified Style Prompt (\textbf{DSP}) technology to diversify style prompts by mixing feature statistics between instance-specific styles. Finally, the generalized visual features are learned through the designed Prompt Modulation (\textbf{PM}) function, which uses visual features as modulation factors. To sum up, the main contributions of this paper are summarized as follows:
\begin{itemize}
	\setlength{\itemsep}{1.0pt}
	\item
	Instead of directly manipulating visual features, it is the first work to explore DG FAS via textual prompt learning, namely CFPL, which allows a broader semantic space to adjust the visual features to generalization. 
	\item
	In order to release the requirement for categories in the text description, our CFPL first learns the prompts conditioned on content and style features with two lightweight transformers, namely Content Q-Former (CQF) and Style Q-Former (SQF). Then, the Prompts are further optimized through two improvements: (1) A Prompt-Text Matched (\textbf{PTM}) aims to ensure CQF learns semantic visual representation; (2) A Diversified Style Prompt (\textbf{DSP}) technology to diversify the learning of style prompts. Finally, the learned prompts modulate visual features to generalization through the designed Prompt Modulation (\textbf{PM}) function.
	\item
	Extensive cross-domain experiments show that the proposed CFPL is effective and outperforms the state-of-the-art (SOTA) methods by an undeniable margin.
\end{itemize}

\section{Related Work}
\subsection{Face Anti-Spoofing}
{\flushleft \textbf{Methods on Intra-datasets.}}
The essence of FAS is a defensive measure for face recognition systems and has been studied for over a decade. Some CNN-based methods~\cite{liu2019deep,liu2020disentangling,Wang_2022_CVPR,huang2022adaptive,liu2022spoof} design a unified framework of feature extraction and classification in an end-to-end manner. Intuitively, the live faces in any scene have consistent face-like geometry. Inspired by this, some works~\cite{Liu2018Learning,shao2020regularized,wang2020deep,yu2020searching} leverage the physical-based depth information instead of binary classification loss as supervision, which are more faithful attack clues in any domain. With the popularity of high-quality 2d attacks, \ie, OULU-NPU~\cite{Boulkenafet2017OULU}, SiW~\cite{Liu2018Learning}, CelebA-Spoof~\cite{CelebA-Spoof} and high-fidelity mask attacks, \ie, MARsV2~\cite{liu20163d2}, WMCA~\cite{george2019biometric,mostaani2020highquality}, HiFiMask~\cite{liu2022contrastive,liu20213d} and SuHiFiMask~\cite{fang2023surveillance,Fang_2023_CVPR} with more realistic in terms of color, texture, and geometry structure, it is very challenging to mine spoofing traces from the visible spectrum alone. Methods based on multimodal fusion~\cite{george2019biometric,zhang2020casia,liu2021casia,liu2021face,george2021cross} have proven to be effective in alleviating the above problems. The motivation for these methods is that indistinguishable fake faces may exhibit quite different properties under the other spectrum. In order to alleviate the limitation of consistency between testing and training modalities, flexible modality based methods~\cite{ijcai2022p165,Liu2023FMViTFM,yu2023flexiblemodal} aims to improve the performance on any single modality by leveraging available multimodal data. However, above methods are not specially designed to solve the domain generalization.

{\flushleft \textbf{Domain Generalization Methods.}}
Domain Adaptation (DA)~\cite{wang2019improving,liu2022source,liu2024source} aims to minimize the distribution discrepancy between the source and target domain by leveraging the unlabeled target data. However, the target data is difficult to collect, or even unknown during training. Domain Generalization (DG) can conquer this by taking the advantage of multiple source domains without seeing any target data. 
MADDG~\cite{Rui2019Multi}, SSDG~\cite{jia2020single}, DR-MD-Net~\cite{Wangarticle} aim to learn a generalized feature space via adversarial training. SSAN~\cite{wang2022domain} reduces the model's overfitting of style by randomly assembling the content and style of the samples. RFM~\cite{shao2020regularized}, MT-FAS~\cite{qin2021meta}, D$^{2}$AM~\cite{chen2021generalizable}, and SDA~\cite{wang2021self} aim to find the generalized feature directions via meta-learning strategies. In addition to aligning a domain-invariant feature space, SA-FAS~\cite{sun2023rethinking} encourages domain separability while aligning the live-to-spoof transition to be the same for all domains. 
Considering the domain information lies in the style features, ANRL~\cite{liu2021adaptive} and SSAN~\cite{wang2022domain} use IN strategy to separate complete representation into content and style features according to image statistics. DRDG~\cite{liu2021dual} iteratively reweights the relative importance between samples to further improve the generalization. The latest emerging strategy is to improve the generalization with the help of domain-specific information. CIFAS~\cite{liu2022causal} adopts causal intervention with backdoor adjustment to mitigate domain bias for learning generalizable features. AMEL~\cite{zhou2022adaptive} exploits the domain-specific feature as a complement to common domain-invariant features to further improve the generalization. IADG~\cite{zhou2023instance} learns generalizable visual features by weakening the features’ sensitivity to instance-specific styles. In addition to supervised training, Liu~et al.~\cite{liu2023towards} propose the first unsupervised DG framework for FAS, which could exploit large amounts unlabeled data to learn generalizable features. 

\subsection{Vision-Language Models (VLMs).}
The vision-language models have undergone a leapfrog development since CLIP~\cite{radford2021learning} was proposed. This approach has stimulated thinking and innovation in many fields, such as object detection, image generation~\cite{liu2024padvg}, and image forgery detection~\cite{liu2023forgery,liu2023unified}. In terms of text models, GPT-3 already has strong language processing capabilities. By combining it with visual basic models to build, and adding some necessary link parameters, the trained vision-language model achieves a combined understanding of images and text. For example, BLIP~\cite{li2023blip} has made significant progress in multimodality by freezing the constructed image encoder and text encoder during training to train an additional small query transformer. Similarly, LLaVa and minigpt-4 reduce the cost of model training by simply linearly mapping image features to the word embedding space. On the basis of deep exploration of VLM by researchers, we want to further extend VLM in FAS. 

\section{CFPL: Class Free Prompt Learning}
\begin{figure*}[t]
	\centering
	\includegraphics[width=0.9\linewidth]{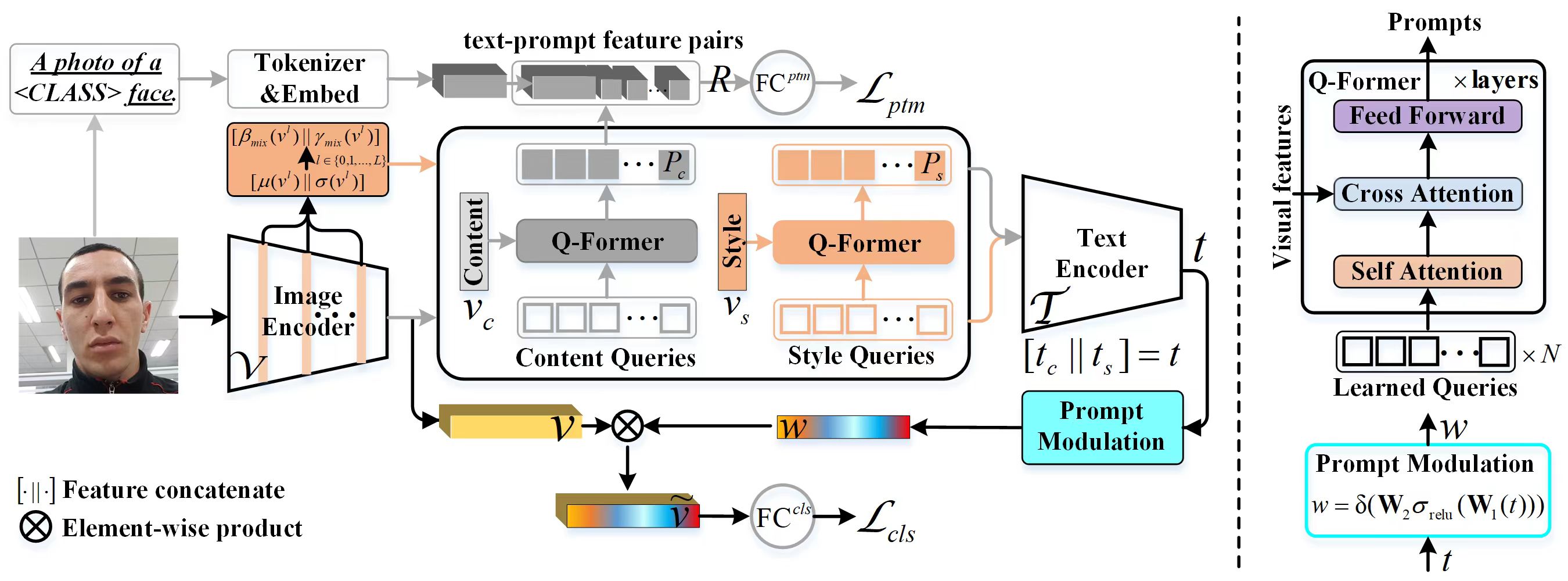}
	\vspace{-0.4cm}
	\caption{Our CFPL is built on CLIP~\cite{radford2021learning} consists of image encoder $\mathcal{V}$ and text encoder $\mathcal{T}$, and adaptes to FAS tasks via prompt learning with four contributions: (1) CQF and SQF. 
	CFPL introduces two lightweight transformers, namely Content Q-Former (CQF) and Style Q-Former (SQF) to learn the different semantic prompts conditioned on content and style features from the image encoder by using a set of learnable query vectors, respectively; (2) Prompt-Text Matched (\textbf{PTM}) surpervision. The fixed template description of each sample is used as a supervise to ensure CQF learns semantic visual representation; (3) A Diversified Style Prompt (\textbf{DSP}). The style from each layer of the image encoder is diversified through mixing feature statistics; 
	(4) Prompt Modulation (\textbf{PM}). The generalized visual feature is adjusted by the modulation factor, which is generated by the text feature through the designed modulation function. 
	}
	\label{fig:architecture}
\end{figure*}

\subsection{Semanticized Prompts Generation}

{\flushleft \textbf{Visual Content and Style features.}}
Considering the challenge of DG FAS is the interference of liveness-irrelevant signals on spoofing cues, we need to model these two types of information with different prompts and alleviate this interference by reducing their correlation. Based on previous research~\cite{wang2022domain,zhou2023instance}, the liveness-irrelevant signals lie more in the instance-specific styles while spoofing clues is an image attribute hidden in the content. Based on this determination, we design Content Q-Former (CQF) and Style Q-Former (SQF) to generate content and style prompts conditioned corresponding visual features, respectively. 
Inspired by Adaptive instance regularization (AdaIN)~\cite{huang2017arbitrary}, given a sample, we first calculate the mean and standard deviation at $l$-th layer, i.e., $\mu(\boldsymbol{v}^{l})$ and $\sigma(\boldsymbol{v}^{l})$; Then, we concatenate them to obtain the style statistics $\boldsymbol{v}^{l}_{s}$ of this layer; Finally, the style feature $\boldsymbol{v}_{s}$ is obtained by averaging style statistics from all layers for this sample. We calculate the content feature $\boldsymbol{v}_{c}$ of the sample from the output of the image encoder by normalizing. The details are as follows:
\begin{equation}\label{Eq:content_style}
	\begin{split}
		&\boldsymbol{v}_{s}=\frac{\sum_{l=1}^{L}\boldsymbol{v}^{l}_{s}}{L}, \boldsymbol{v}^{l}_{s} =\left[\mu(\boldsymbol{v}^{l})\left|\right|\sigma(\boldsymbol{v}^{l})\right], \boldsymbol{v}_{s}\in \mathbb{R}^{1\times 2d}, \\
		&\boldsymbol{v}_{c}=\frac{\boldsymbol{v}^{L}-\mu(\boldsymbol{v}^{L})}{\sigma(\boldsymbol{v}^{L})}, \boldsymbol{v}_{c}\in \mathbb{R}^{n\times d}
	\end{split}
\end{equation}
where $L$ is the total layers of the image encoder. $\left[\cdot\left|\right|\cdot\right]$ represents concatenating features along embedding dimension.

{\flushleft \textbf{CQF and SQF.}}
As shown in Fig.~\ref{fig:architecture}, CQF and SQF share a similar backbone, which consists of alternating layers of multiheaded self-attention ($\texttt{MSA}$), multiheaded cross-attention ($\texttt{MCA}$) and $\texttt{MLP}$ blocks. Firstly, we create $N$ learnable query embeddings $\boldsymbol{Q}=\left \{\boldsymbol{q}^{1}, \boldsymbol{q}^{2},\cdots, \boldsymbol{q}^{N}\right \}\in \mathbb{R}^{N\times d}$ as input to the backbone, where each query has a dimension of $d=512$ (same dimension with multi-modal embedding space); Then, the queries interact with each other through MSA block, and interact with image features $\boldsymbol{v}\in \mathbb{R}^{d}$ through MCA block; Finally, we obtain the prompt $\boldsymbol{P}=\left \{\boldsymbol{p}^{1}, \boldsymbol{p}^{2},\dots,\boldsymbol{p}^{N}  \right \} \in \mathbb{R}^{N\times d}$ after the queries pass through MLP block. This process can be expressed as:
\begin{equation}\label{Eq:tpl}
	\begin{split}	
		\boldsymbol{Q'}&=\boldsymbol{Q}+\texttt{MSA}(\texttt{LN}(\boldsymbol{Q})), \boldsymbol{Q'}\in \mathbb{R}^{N\times d}\\
		\boldsymbol{Q''}&=\boldsymbol{Q'}+\texttt{MCA}(\texttt{LN}(\boldsymbol{Q'}), \texttt{LN}(\boldsymbol{v})),\boldsymbol{Q''}\in \mathbb{R}^{N\times d} \\
		\boldsymbol{P}&=\boldsymbol{Q''}+\texttt{MLP}(\texttt{LN}(\boldsymbol{Q''})),\boldsymbol{P}\in \mathbb{R}^{N\times d}
	\end{split}
\end{equation}
where Layernorm ($\texttt{LN}$) is applied before every block, and residual connections after every block. The $\texttt{MLP}$ contains two layers with a $\texttt{GELU}$ non-linearity. Based on this training mechanism, CQF and SQF can bridge the gap between visual and language modalities by interacting prompt queries with corresponding image features, and output the content prompt $\boldsymbol{P}_{c}$ and style prompt $\boldsymbol{P}_{s}$, respectively.

\subsection{Generalized Prompt Optimization}
{\flushleft \textbf{Text Supervision in Content Prompt.}}

Due to the lack of semantics for CLIP in the FAS categories, it is not suitable to align queries and text representations with the concept of maximizing their mutual information. Instead, we guide the CQF to understand the FAS's categories at a higher level with a binary classification task, where the model is asked to predict whether a prompt-text pair is matched (PTM). 

Given a min-batch with $B$ samples, we first generate a text description containing category attribute for each sample, such as ``a photo of a $\left\langle\mathrm{CLASS}\right\rangle$ face.", where ``$\mathrm{CLASS}$" is ``live" for real face and ``fake" for spoof face, respectively; After that, the text descriptions $\boldsymbol{T}$ are transformed into text supervisions $\boldsymbol{S}\in \mathbb{R}^{B\times 77\times d}$ through $\texttt{Tokenizer}$ and $\texttt{Embed}$ layers, sequentially; Then, we construct positive and negative feature pairs of prompt-text for prediction by CQF. Specifically, we concatenate content prompt $\boldsymbol{P_{c}} \in \mathbb{R}^{B\times N\times d}$ and text supervision $\boldsymbol{S}$ according to the embedding dimension, and obtain the positive feature pairs $\boldsymbol{R}_{p} \in \mathbb{R}^{B\times N\times 2d}$. We adopt the hard negative mining strategy from ALBEF~\cite{li2021align} to create informative negative pairs. Such as for each prompt, one negative text is selected with the contrastive similarity distribution, where texts that are more similar to the prompt have a higher chance of being sampled. 
A similar strategy for one hard negative prompt for each text. Therefore, we can obtain negative feature pairs $\boldsymbol{R}^{prompt}_{n} \in \mathbb{R}^{B\times N\times 2d}$ and $\boldsymbol{R}^{text}_{n} \in \mathbb{R}^{B\times N\times 2d}$ by mining prompt and text, respectively; After that, we concatenate all positive and negative feature pairs $\boldsymbol{R}_{p}$, $\boldsymbol{R}^{prompt}_{n}$, and $\boldsymbol{R}^{text}_{n}$ to obtain the joint features $\boldsymbol{R} \in \mathbb{R}^{3B\times N\times 2d}$ according to the batch dimension. This process can be expressed as:
\begin{equation}\label{Eq:R}
	\begin{split}	
		\boldsymbol{S}&=\texttt{Embed}(\texttt{Tokenizer}(\boldsymbol{T})), \boldsymbol{S}\in \mathbb{R}^{{B\times 77\times d}}, \\
		\boldsymbol{S}&=\texttt{Mean\&Expand}(\boldsymbol{S}), \boldsymbol{S}\in \mathbb{R}^{{B\times N\times d}}, \\
		\boldsymbol{R}_{p}&=\left [\boldsymbol{P}\left|\right|\boldsymbol{S}\right]_{\texttt{2}}, \boldsymbol{R}_{p}\in \mathbb{R}^{{B\times N\times 2d}},\\
		\boldsymbol{R}&=\left [ \boldsymbol{R}_{p}\left|\right|\boldsymbol{R}^{prompt}_{n}\left|\right|\boldsymbol{R}^{text}_{n}\right ]_{\texttt{0}}, \boldsymbol{R}\in \mathbb{R}^{{3B\times N\times 2d}}\\
	\end{split}
\end{equation}
where the number of word tokens in text supervisions is aligned with the number of queries in the content prompt by averaging ($\texttt{Mean}$) and expanding ($\texttt{Expand}$) $N$ times. $\left[\cdot\left|\right|\cdot\right]_{\texttt{dim}}$ represents concatenating features along $\texttt{dim}$ dimension. Finally, the optimization of text supervision is achieved by predicting the matched and unmatched probabilities for the joint features $\boldsymbol{R}$:
\begin{equation}\label{Eq:ptm}
	\begin{split}	
		\mathcal{L}_{ptm}=\sum_{i=1}^{3B} \mathcal{H}(\boldsymbol{y}^{ptm}_{i},\texttt{Mean}(\texttt{FC}^{ptm}(\boldsymbol{R}_{i})))
	\end{split}
\end{equation}
where we feed each query embedding into a two-class linear classifier to obtain a logit, and average ($\texttt{Mean}$) the logits across all queries as the output matching score. $\mathcal{H}(.,.)$ is the cross-entropy loss, $\texttt{FC}^{ptm}$ is a fully-connected layer followed by softmax, and $\boldsymbol{y}^{ptm}\in\left\{0,1\right\}$ is a 2-dimensional one-hot vector representing the ground-truth label.

\begin{table*}[]
	\centering
	\scalebox{0.80}{
		\begin{tabular}{lccccccccccccc}
			\hline
			\multirow{3}{*}{Method} & \multicolumn{3}{c}{OCI→M}                                                                                              & \multicolumn{3}{c}{OMI→C}                                                                                              & \multicolumn{3}{c}{OCM→I}                                                                                              & \multicolumn{3}{c}{ICM→O}                                                                                              & avg.                  \\ \cline{2-14} 
			& \multirow{2}{*}{HTER$\downarrow$} & \multirow{2}{*}{AUC } & \multirow{2}{*}{\begin{tabular}[c]{@{}c@{}}TPR@\\ FPR=1\% \end{tabular}} & \multirow{2}{*}{HTER} & \multirow{2}{*}{AUC } & \multirow{2}{*}{\begin{tabular}[c]{@{}c@{}}TPR@\\ FPR=1\% \end{tabular}} & \multirow{2}{*}{HTER} & \multirow{2}{*}{AUC } & \multirow{2}{*}{\begin{tabular}[c]{@{}c@{}}TPR@\\ FPR=1\% \end{tabular}} & \multirow{2}{*}{HTER} & \multirow{2}{*}{AUC } & \multirow{2}{*}{\begin{tabular}[c]{@{}c@{}}TPR@\\ FPR=1\%\end{tabular}} & \multirow{2}{*}{HTER} \\
			&                       &                      &                                                                         &                       &                      &                                                                         &                       &                      &                                                                         &                       &                      &                                                                         &                       \\ \hline
			MADDG~\cite{Rui2019Multi}                   & 17.69                 & 88.06                & -                                                                       & 24.50                 & 84.51                & -                                                                       & 22.19                 & 84.99                & -                                                                       & 27.98                 & 80.02                & -                                                                       & 23.09                 \\
			DR-MD-Net~\cite{Wangarticle}               & 17.02                 & 90.10                & -                                                                       & 19.68                 & 87.43                & -                                                                       & 20.87                 & 86.72                & -                                                                       & 25.02                 & 81.47                & -                                                                       & 20.64                 \\
			RFMeta~\cite{shao2020regularized}    & 13.89    & 93.98  & -                                                                       & 20.27                 & 88.16                & -                                                                       & 17.30                 & 90.48                & -                                                                       & 16.45                 & 91.16                & -                                                                       & 16.97                 \\
			NAS-FAS~\cite{yu2020nasfas}                 & 19.53                 & 88.63                & -                                                                       & 16.54                 & 90.18                & -                                                                       & 14.51                 & 93.84                & -                                                                       & 13.80                 & 93.43                & -                                                                       & 16.09                 \\
			D$^{2}$AM~\cite{chen2021generalizable}                    & 12.70                 & 95.66                & -                                                                       & 20.98                 & 85.58                & -                                                                       & 15.43                 & 91.22                & -                                                                       & 15.27                 & 90.87                & -                                                                       & 16.09                 \\
			SDA~\cite{wang2021self}                     & 15.40                 & 91.80                & -                                                                       & 24.50                 & 84.40                & -                                                                       & 15.60                 & 90.10                & -                                                                       & 23.10                 & 84.30                & -                                                                       & 19.65                 \\
			DRDG~\cite{liu2021dual}                    & 12.43                 & 95.81                & -                                                                       & 19.05                 & 88.79                & -                                                                       & 15.56                 & 91.79                & -                                                                       & 15.63                 & 91.75                & -                                                                       & 15.66                 \\
			ANRL~\cite{liu2021adaptive}                    & 10.83                 & 96.75                & -                                                                       & 17.83                 & 89.26                & -                                                                       & 16.03                 & 91.04                & -                                                                       & 15.67                 & 91.90                & -                                                                       & 15.09                 \\
			SSDG-R~\cite{jia2020single}                  & 7.38                  & 97.17                & -                                                                       & 10.44                 & 95.94                & -                                                                       & 11.71                 & 96.59                & -                                                                       & 15.61                 & 91.54                & -                                                                       & 11.28                 \\
			SSAN-R~\cite{wang2022domain}                  & 6.67                  & 98.75                & -                                                                       & 10.00                 & 96.67                & -                                                                       & 8.88                  & 96.79                & -                                                                       & 13.72                 & 93.63                & -                                                                       & 9.81                  \\
			PatchNet~\cite{Wang_2022_CVPR}                & 7.10                  & 98.46                & -                                                                       & 11.33                 & 94.58                & -                                                                       & 13.40                 & 95.67                & -                                                                       & 11.82                 & 95.07                & -                                                                       & 10.91                 \\
			SA-FAS~\cite{sun2023rethinking}                  & 5.95                  & 96.55                & -                                                                       & 8.78                  & 95.37                & -                                                                       & 6.58                  & 97.54                & -                                                                       & 10.00                 & 96.23                & -                                                                       & 7.82                  \\
			IADG~\cite{zhou2023instance}                    & 5.41                  & 98.19                & -                                                                       & 8.70                  & 96.44                & -                                                                       & 10.62                 & 94.50                & -                                                                       & 8.86                  & 97.14                & -                                                                       & 8.39                  \\ 
			CFPL(Ours)      & \textbf{3.09}                  & \textbf{99.45}                & 94.28   & \textbf{2.56}     & \textbf{99.10}       & 66.33                                                                   & \textbf{5.43}                  & \textbf{98.41}                & 85.29                                                                   & \textbf{3.33}                  & \textbf{99.05}                & 90.06       & \textbf{3.60}                  \\ \hline
			ViTAF*-5-shot~\cite{huang2022adaptive}           & 2.92                  & 99.62                & 91.66                                                                   & 1.40                  & 99.92                & 98.57                                                                   & 1.64                  & 99.64                & 91.53                                                                   & 5.39                  & 98.67                & 76.05                                                                   & 2.83                  \\
			FLIP-MCL*~\cite{srivatsan2023flip}               & 4.95                  & 98.11                & 74.67                                                                   & 0.54                  & 99.98                & 100.00                                                                  & 4.25                  & 99.07                & 84.62                                                                   & 2.31                  & 99.63                & 92.28                                                                   & 3.01                  \\
			CFPL*(Ours)      & 1.43                  & 99.28         & 98.57   & 2.56     & 99.10       & 66.33                                                           & 5.43                  & 98.41                & 85.29        & 2.50      & 99.42               & 94.72       & 2.98                    \\ \hline
		\end{tabular}
	}
	\caption{The results (\%) of Protocol 1 on MSU-MFSD (M), CASIA-FASD (C), ReplayAttack (I), and OULU-NPU (O) datasets. Note that the $*$ indicates the corresponding method using CelebA-Spoof~\cite{CelebA-Spoof} as the supplementary source dataset and `5-shot' represents 5 images from the target datasets participating in the training phase.}
	\label{tab:MCIO_results}
\end{table*}

{\flushleft \textbf{Diversified Style Prompt.}}
Due to the indescribability of the sample style, we are unable to complete this task using text supervision. Implicitly, we borrow a strategy from MixStyle~\cite{zhou2021domain} that mixes style feature statistics between instances to achieve diversification of style prompts.

Specifically, given the visual style stastics $\left[\mu(\boldsymbol{v}), \sigma(\boldsymbol{v}) \right]$ of a min-batch, we first obtain reference statistics $\mu(\hat{\boldsymbol{v}})$ and $\sigma(\hat{\boldsymbol{v}})$ by shuffling the order of batch dimension for $\mu(\boldsymbol{v})$ and $\sigma(\boldsymbol{v})$, respectively; Then, we generate mixture of feature statistics $\gamma_{mix}$ and $\beta_{mix}$ through a weighted approach:
\begin{equation}\label{Eq:MixStyle}
	\begin{split}		
		\gamma_{mix}=\lambda \sigma(\boldsymbol{v})+(1-\lambda)\sigma(\hat{\boldsymbol{v}}), \\
		\beta_{mix}=\lambda \mu(\boldsymbol{v})+(1-\lambda)\mu(\hat{\boldsymbol{v}})
	\end{split}
\end{equation}
where $\lambda$ is an instance-specific, random weight sampled from the beta distribution, $\lambda\sim Beta(\alpha, \alpha)$. $\alpha$ is set to 0.1 according to the suggestion in~\cite{zhou2021domain}. Finally, the mixture of style statistics $\left[\beta_{mix}, \gamma_{mix} \right]$ are used to calculate style features according to Eq.~\ref{Eq:content_style}.

\subsection{Prompt Modulation on Visual Features}
{\flushleft \textbf{Class Free Prompt Modulation.}}
Due to the content and style prompts are generated based on sample instances, they are more suitable as a set of fine-tuning factors (class free) for adaptively recalibrating channel-wise visual feature responses, compared to using them as classifier's weights (with class) to predict visual feature. In this work, we use prompts to distinguish between generalized features and liveness-irrelevant signals by explicitly modeling interdependencies between channels.

Concretely, we first input content prompt $\boldsymbol{P_{c}}$ and style prompt $\boldsymbol{P_{s}}$ into the text encoder to produce text features, $\boldsymbol{t_{c}}\in \mathbb{R}^{B\times d}$ and $\boldsymbol{t_{s}}\in \mathbb{R}^{B\times d}$, respectively; After that, we concatenate these two types of text features along the embedding dimension to obtain modulation features $\boldsymbol{t}\in \mathbb{R}^{B\times 2d}$ with rich visual concepts; Then, we employ a gating mechanism $\texttt{g}_{e}$ with a sigmoid activation $\delta$ to map modulation features to the weighting factors $\boldsymbol{w}\in \mathbb{R}^{B\times d}$. This process can be expressed as:
\begin{equation}\label{Eq:senet}
	\begin{split}		
		\boldsymbol{w}&=\delta(\texttt{g}_{e}(\boldsymbol{t},\mathbf{W}))=\delta(\mathbf{W}_{2}\sigma_{\textrm{relu}}(\mathbf{W}_{1}\boldsymbol{t})), \\
		\boldsymbol{\tilde{v}}&=\left [\boldsymbol{\tilde{v}}^{1}, \boldsymbol{\tilde{v}}^{2},\dots , \boldsymbol{\tilde{v}}^{d} \right], \boldsymbol{\tilde{v}}^{c}=\boldsymbol{w}^{c}\cdot \boldsymbol{v}^{c}, \\
		\mathcal{L}_{cls}&=\sum_{i=1}^{B} \mathcal{H}(\boldsymbol{y}^{cls}_{i},\texttt{FC}^{cls}(\boldsymbol{\tilde{v}}_{i})), \boldsymbol{\tilde{v}}\in \mathbb{R}^{B\times d}
	\end{split}
\end{equation}
where $\sigma_{\textrm{relu}}$ is the ReLU function, $\mathbf{W}_{1}\in \mathbb{R}^{\frac{d}{r}\times 2d}$ and $\mathbf{W}_{2}\in \mathbb{R}^{d\times \frac{d}{r}}$ are trainable parameters for two fully connected (FC) layers in $\texttt{g}_{e}$ function. $r$ is a reduction ratio, with a value of 16 in this work. Finally, the adapted visual features $\boldsymbol{\tilde{v}}\in \mathbb{R}^{B\times d}$ are obtained by weighting the channel-wise feature $\boldsymbol{v}^{c}$ with the scalar $\boldsymbol{w}^{c}$, and append a fully-connected ($\texttt{FC}^{cls}$) layer followed by softmax to predict a two-class (i.e., live or fake) probability. $y^{cls}\in\left\{0,1\right\}$ is the label for live or spoof face.

{\flushleft \textbf{Model Training and Inference.}}
In the training stage, parameters from two designed Q-Formers, i.e., CQF and SQF, two fully connected layers for classifiers, i.e., $\texttt{FC}^{ptm}$ and $\texttt{FC}^{cls}$, one gate function, i.e., $\texttt{g}_{e}$, and image encoder $\mathcal{V}$ are updated and the text encoder $\mathcal{T}$ is fixed. The full training objective of CFPL is:
\begin{equation}\label{Eq:loss}
	\begin{split}		
		\mathcal{L}_{total}=\mathcal{L}_{cls}+\mathcal{L}_{ptm}
	\end{split}
\end{equation}

In the inference stage, 
our CQF and SQF will adaptively generate the semanticized prompt as input to the text encoder based on each sample instance. Finally, the text encoder generates continuous and widely adjustable modulation factors for weighting visual features to generalization.

\begin{table*}[]
	\centering
	\scalebox{1.0}{
		\begin{tabular}{lcccccccccc}
			\hline
			\multirow{3}{*}{Method} & \multicolumn{3}{c}{CS→W}                                                                                               & \multicolumn{3}{c}{SW→C}                                                                                               & \multicolumn{3}{c}{CW→S}                                                                                               & avg.                  \\ \cline{2-11} 
			& \multirow{2}{*}{HTER$\downarrow$} & \multirow{2}{*}{AUC } & \multirow{2}{*}{\begin{tabular}[c]{@{}c@{}}TPR@\\ FPR=1\% \end{tabular}} & \multirow{2}{*}{HTER} & \multirow{2}{*}{AUC } & \multirow{2}{*}{\begin{tabular}[c]{@{}c@{}}TPR@\\ FPR=1\% \end{tabular}} & \multirow{2}{*}{HTER} & \multirow{2}{*}{AUC } & \multirow{2}{*}{\begin{tabular}[c]{@{}c@{}}TPR@\\ FPR=1\% \end{tabular}} & \multirow{2}{*}{HTER} \\
			&                       &                      &                                                                         &                       &                      &                                                                         &                       &                      &                                                                         &                       \\ \hline
			ViT*~\cite{huang2022adaptive}                    & 7.98                  & 97.97                & 73.61                                                                   & 11.13                 & 95.46                & 47.59                                                                   & 13.35                 & 94.13                & 49.97                                                                   & 10.82                 \\
			ViTAF*-5-shot~\cite{huang2022adaptive}           & 2.91                  & 99.71                & 92.65                                                                   & 6.00                  & 98.55                & 78.56                                                                   & 11.60                 & 95.03                & 60.12                                                                   & 6.83                  \\
			FLIP-MCL*~\cite{srivatsan2023flip}               & 4.46                  & 99.16                & 83.86                                                                   & 9.66                  & 96.69                & 59.00                                                                   & 11.71                 & 95.21                & 57.98                                                                   & 8.61                  \\ 
			CFPL*(Ours)   & 4.40       & 99.11     & 85.23  & 8.13     & 96.70  & 62.41   & 8.50  & 97.00                    & 55.66                                                                      & 7.01                     \\ 
			\hline
			ViT~\cite{huang2022adaptive}                     & 21.04                 & 89.12                & \textbf{30.09}                                                                   & 17.12                 & 89.05                & 22.71                                                                   & 17.16                 & 90.25                & 30.23                                                                   & 18.44                 \\
			CLIP-V~\cite{radford2021learning}                  & 20.00                 & 87.72                & 16.44                                                                   & 17.67                 & 89.67                & 20.70                                                                   & \textbf{8.32}                  & \textbf{97.23}                & 57.28                                                                   & 15.33                 \\
			CLIP~\cite{radford2021learning}                  & 17.05                 & 89.37             & 8.17     & 15.22                 & \textbf{91.99}                & 17.08               & 9.34                  & 96.62                &\textbf{60.75}                & 13.87                 \\
			CoOp~\cite{zhou2022coop}                    & 9.52                  & 90.49                & 10.68                                                                   & 18.30                 & 87.47                & 11.50                                                                   & 11.37                 & 95.46                & 40.40                                                                   & 13.06                 \\
			CFPL (Ours)              & \textbf{9.04}      & \textbf{96.48}                & 25.84                                                                   & \textbf{14.83}                 & 90.36                & \textbf{8.33}                                                                    & 8.77                  & 96.83                & 53.34                                                                   & \textbf{10.88}                 \\
			\hline
		\end{tabular}
	}
	\caption{The results (\%) of Protocol 2 on CASIA-SURF (S), CASIA-SURF CeFA (C), and WMCA (W) datasets. Note that the $*$ indicates the corresponding method using CelebA-Spoof~\cite{CelebA-Spoof} as the supplementary source dataset and `5-shot' represents 5 images from the target datasets participating in the training phase.}
	\label{tab:SCW_results}
\end{table*}

\section{Experimental Setup}
{\flushleft \textbf{Datasets, Protocols and Evaluation Metrics.}}
Following the prior work~\cite{huang2022adaptive}, two Protocols are used to evaluate the generalization in this work. For Protocol 1, we use four widely-used benchmark datasets, MSU-MFSD (M)~\cite{wen2015face}, CASIA-FASD (C)~\cite{Zhang2012A}, Idiap Replay-Attack (I)~\cite{Chingovska_BIOSIG-2012}, and OULU-NPU (O)~\cite{Boulkenafet2017OULU}. For Protocol 2, we use RGB samples in CASIA-SURF (S)~\cite{zhang2020casia}, CASIA-SURF CeFA (C)~\cite{liu2021casia}, and WMCA (W)~\cite{george2019biometric} datasets, which contain more subjects, diverse attack types, and rich collection environments. In each Protocol, we treat each dataset as a domain and apply the leave-one-out testing for generalization evaluation. We adopt three metrics to evaluate the performance of a model: (1) HTER. It computes the average of the FRR and the FAR. (2) AUC. It evaluates the theoretical performance of the model. (3) TPR at a fixed False Positive Rate (FPR=1\%). It can be used to select a suitable trade-off threshold according to a given real application.

{\textbf{Implementation Details.}}
We set the length of style and content queries to $16$, where each query has a dimension of $512$; The depth of the CQF and SQF is set to $1$. 
Style prompt diversification is activated in the training phase with a probability of 0.5 and does not participate in the test phase; All models are trained with a batch size of $12$, an Adam optimizer with a weight decay of 0.05. 
The minimum learning rate at the second stage is $1e-6$. We resize images to $224\times224$, augmented with random resized cropping and horizontal flipping, and train all models with $500$ epochs. 

\subsection{Cross-domain Results}
For Protocol 1, we report the results of recent SOTA methods on Tab.~\ref{tab:MCIO_results}, such as SA-FAS~\cite{sun2023rethinking} encourages domain separability while aligning the live-to-spoof transition; IADG~\cite{zhou2023instance} learns the generalizable feature by weakening the features’ sensitivity to instance-specific styles; FLIP-MCL*~\cite{srivatsan2023flip} improves the visual representations with the help of natural language, and ViTAF*-5-shot~\cite{huang2022adaptive} tackles a real-world anti-spoofing when $5$ images are available from target datasets. 
From Tab.~\ref{tab:MCIO_results}, without using CelebA-Spoof~\cite{CelebA-Spoof}, it can be seen that our method achieves the best performance for all metrics on four test datasets. Specifically, on the HTER metric (similar conclusions on AUC), CFPL outperforms IADG~\cite{zhou2023instance} for all target domains with lower values: \textbf{M} (3.09\% \vs, 5.41\%), \textbf{C} (2.56\% \vs, 8.70\%), \textbf{I} (5.43\% \vs, 10.62\%) and \textbf{O} (3.33\% \vs, 8.86\%). Finally, an average HTER of 3.60\% is achieved, significantly better than the previous best result of 7.82\%. 
After introducing the CelebA-Spoof~\cite{CelebA-Spoof} dataset as the supplementary training data, CFPL* outperforms FLIP-MCL*~\cite{srivatsan2023flip} in terms of the average metric of HTER, such as 2.98\% \vs, 3.01\%. It indicates that our algorithm aligns images with text representation by class-free prompt learning is superior to an ensemble of class descriptions. In addition, CFPL* is slightly inferior to the ViTAF*-5-shot~\cite{huang2022adaptive} on the average of HTER, such as 2.98\% \vs, 2.83\%, which uses 5 additional samples from target domain in training data, greatly alleviating the overfitting on the domain-specific distribution.

For Protocol 2, we list the results of different methods on Tab.~\ref{tab:SCW_results}. Without using CelebA-Spoof~\cite{CelebA-Spoof}, our CFPL significantly surpasses several baselines in terms of the HTER metric when tested on the \textbf{W} and \textbf{C} domains. 
However, when tested on the \textbf{S} dataset, CFPL is slightly inferior to the CLIP-V~\cite{radford2021learning} on the HTER (8.77\% \vs 8.32\%), AUC (96.83\% \vs 97.23\%) and TPR@FPR=1\% (53.34\% \vs 57.28\%) metrics. Due to the obvious spoofing traces on the CASIA-SURF dataset~\cite{zhang2020casia}, relying solely on visual features is relatively generalizable. Similar to the conclusion of Protocol 1, CFPL* outperforms baseline ViT*~\cite{huang2022adaptive} and FLIP-MCL*~\cite{srivatsan2023flip} when introducing the CelebA-Spoof~\cite{CelebA-Spoof} dataset into training data. For example, our CFPL* achieves the optimal mean on the HTER (7.01\%) metric.
\begin{table}[]
	\centering
	\scalebox{0.82}{
		\begin{tabular}{@{}cccc|c|c|c@{}}
			\toprule
			Baseline & PTM & DSP & PM & HTER(\%)$\downarrow$ & AUC(\%)  & \begin{tabular}[c]{@{}c@{}}TPR(\%)\\ @FPR=1\% \end{tabular} \\ \midrule
			CoOp~\cite{zhou2022coop}& -   & -   & -  & 8.78     & 94.77   &  43.71  \\
			$\checkmark$ & -   & -   & -  & 8.11     & 96.09   & 51.59   \\
			$\checkmark$ & $\checkmark$   & -   & -  & 7.50     & 96.39   & 54.78 \\
			$\checkmark$ & $\checkmark$   & $\checkmark$   & -  & 7.08     & 96.79   & 57.61 \\
			$\checkmark$ & $\checkmark$   & $\checkmark$   & $\checkmark$  & 6.72     & 97.09   & 60.35  \\ \bottomrule
		\end{tabular}
	}
	\caption{Ablation of each component, where each result is the average on all sub-protocols.}
	\label{tab:Ablation_component}
\end{table}

\subsection{Ablation Study}
{\textbf{Effectiveness of each component.}}
In order to evaluate the contribution of each component in our framework, such as Text Supervision (abbreviated as `PTM'), Diversification of Style Prompt (abbreviated as `DSP'), and Prompt Modulation (abbreviated as `PM'), we conduct ablation studies on Protocol 1 and 2 by gradually introducing one of them into the Baseline (abbreviated as `B'), where the Baseline is obtained by removing all contributions from the CFPL. And report the average results on all sub-protocols in Tab.~\ref{tab:Ablation_component}. 

Specifically, instead of modeling a prompt’s context words with free learnable vectors in CoOp~\cite{zhou2022coop}, our `Baseline' introduces two lightweight transformers CQF and SQF to learn the expected prompts conditioned on content and style features from the image encoder, achieving significant generalization benefits of -0.67\% (HTER), +1.32\% (AUC) and +7.88\% (TPR@FPR=1\%), respectively. After introducing the text supervision in the content prompt, the results of the three metrics are optimized to 7.50\% (HTER), 96.39\% (AUC), and 54.78\% (TPR@FPR=1\%), respectively. It indicates that promoting the content prompt carries sufficient category attributes that can be converted into generalization benefits. Further diversification of style statistics in style prompts can further expand performance benefits, i.e., -0.42\% (HTER), +0.4\% (AUC), and +2.83\% (TPR@FPR=1\%). Instead of using the designed gate function $\texttt{g}_{e}$, we calculate the modulation factor $\boldsymbol{w}$ by calculating the mean of content and style features, such as $\boldsymbol{w}=\left (\boldsymbol{t_{c}}+\boldsymbol{t_{s}} \right)/2, \boldsymbol{w}\in \mathbb{R}^{{B\times d}}$. From the Tab.~\ref{tab:Ablation_component}, it can be seen that if removing the designed gate function, the generalization significantly decreased from 6.72\% (HTER), 97.09\% (AUC) and 60.35\% (TPR@FPR=1\%) to 7.08\% (HTER), 96.79\% (AUC) and 57.61\% (TPR@FPR=1\%). 

\begin{figure*}[t]
	\centering
	\begin{minipage}{0.3\linewidth}
		\centerline{\includegraphics[width=1\textwidth]{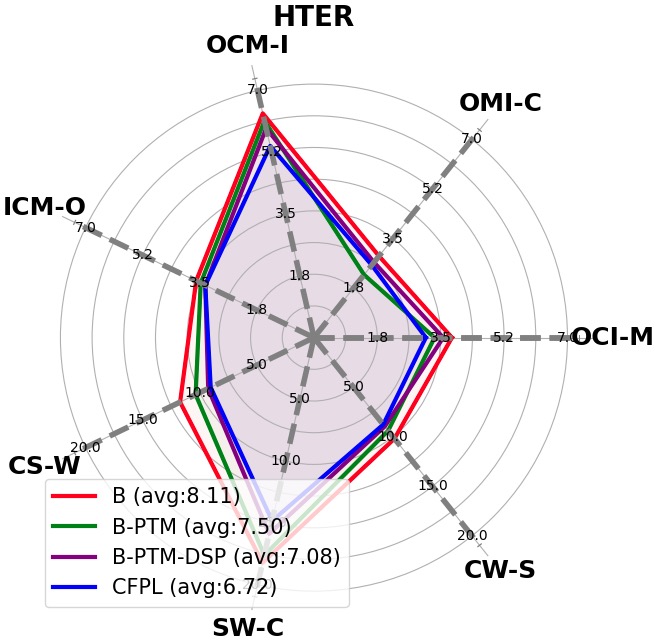}}
		\centerline{(a)}
	\end{minipage}
	\
	\begin{minipage}{0.3\linewidth}
		\centerline{\includegraphics[width=1\textwidth]{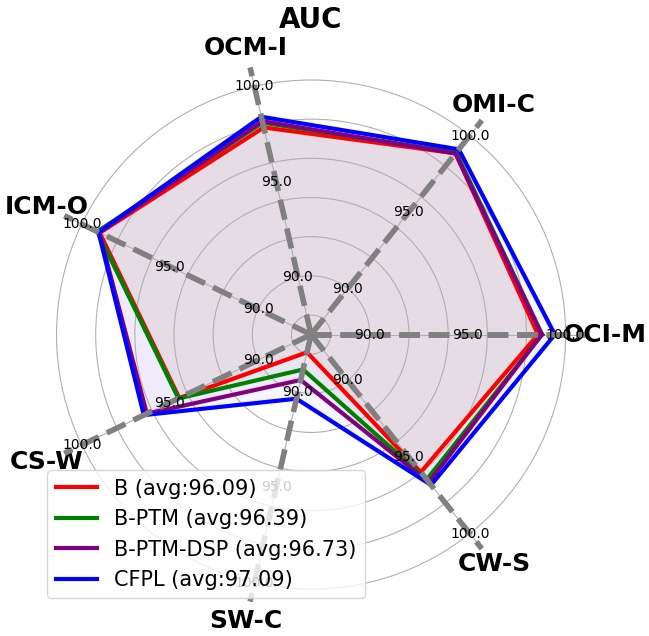}}
		\centerline{(b)}
	\end{minipage}
	\
	\begin{minipage}{0.3\linewidth}
		\centerline{\includegraphics[width=1\textwidth]{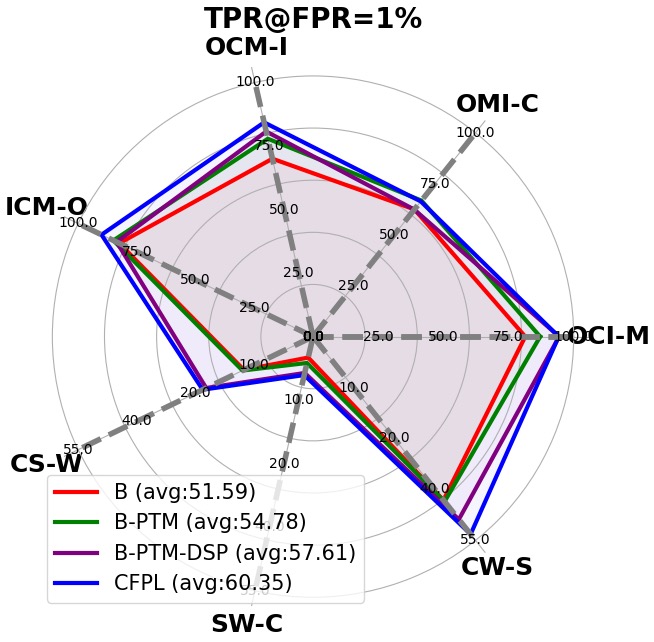}}
		\centerline{(c)}
	\end{minipage}
	\vspace{-0.2cm}
	\caption{The results of each method on three metrics across all sub-protocols, where the red line represents the Baseline, and the blue line represents our CFPL. For the HTER metric, the smaller area enclosed by lines, the better performance of the corresponding methods. The opposite conclusion applies to metrics AUC and TPR@FPR=1\%.}
	\label{fig:Ablation_component}
\end{figure*}

Furthermore, in Fig.~\ref{fig:Ablation_component}, we detailed the result of each method on three metrics across all sub-protocols, where the red line represents the Baseline, and the blue line represents our CFPL. From Fig.~\ref{fig:Ablation_component}, it can be clearly seen that the blue line shrinks with the smallest area in the polar coordinate system of the HTER, while is distributed with the largest area for the AUC and TPR@FPR=1\%. The opposite conclusion applies to the Baseline of red lines. Almost all methods have an undeniable advantage over Baseline. 

{\textbf{Effect of the Structures of CQF and SQF.}}
In Tab.~\ref{tab:Ablation_structures}, we list the results of CoCoOp~\cite{zhou2022cocoop}, CFPL removing SQF (abbreviated as `CQF'), CFPL removing CQF (abbreviated as `SQF'), and CFPL on the ICM→O experiment. From Tab.~\ref{tab:Ablation_structures}, we can see that a simple two-layer bottleneck structure cannot effectively alleviate the FAS task with significant domain differences, such as achieving results of 6.80\%, 97.27\%, and 60.41\% on metrics HTER, AUC, and TPR@FPR=1\%. When replacing Meta-Net with CQF and SQF, we obtain performance benefits of -1.68\% (HTER), +1.38\% (AUC), and +13.26\% (TPR@FPR=1\%) and -1.96\% (HTER), +1.48\% (AUC) and +26.67\% (TPR@FPR=1\%), respectively. Finally, by collaborating with CQF and SQF, our CFPL achieves the best results, such as 3.33\% (HTER), 99.05\% (AUC), and 90.06\% (TPR@FPR=1\%). Furthermore, our CFPL further improves their performance through collaborative CQF and SQF,  indicating that the CQF and SQF not only bring significant benefits but also have a positive collaborative effect.
\begin{table}[]
	\centering
	\scalebox{0.92}{
		\begin{tabular}{@{}c|ccc@{}}
			\toprule
			Method   & \multicolumn{1}{c|}{{HTER(\%)$\downarrow$}} & \multicolumn{1}{c|}{AUC(\%) } & TPR(\%)@FPR=1\%  \\ \midrule
			CoCoOp~\cite{zhou2022cocoop}   & 6.80         & 97.27       & 60.41    \\ \midrule
			CQF      & 5.12    & 98.65   & 73.67   \\
			SQF      & 4.84   & 98.75    & 87.08   \\ \midrule
			CFPL & 3.33    & 99.05   & 90.06               \\ \bottomrule
		\end{tabular}
	}
	\caption{Ablation of the structures for CQF and SQF on ICM→O.}
	\label{tab:Ablation_structures}
\end{table}
\begin{table}[]
	\centering
	\scalebox{1.1}{
		\begin{tabular}{@{}c|cccc@{}}
			\toprule
			HTER(\%)$\downarrow$  & \multicolumn{4}{c}{Length} \\ \midrule
			Depth    & $\times$8   & $\times$16   & $\times$32  & $\times$64  \\ \midrule
			$\times$1                & 3.47                  & \textit{3.33}               & \textit{3.33}            & \textit{\textbf{3.30}}    \\
			$\times$4                & 3.45                 & \textbf{3.42}             & 3.45    & 3.45                   \\
			$\times$8                & 3.56                 & 3.56                          & \textbf{3.47}          & 3.47    \\
			$\times$12               & \textit{3.41}       & \textbf{3.33}             & 3.33                      & 3.33   \\ \bottomrule
		\end{tabular}
	}
	\caption{Ablation of the length for Queries and the depth for Q-former on ICM→O. The optimal value for each row/column is represented in bold/italics.}
	\label{tab:Ablation_depth_length}
\end{table}

{\textbf{The Length of Queries and the Depth of Q-Former.}}
The number of learnable queries and the depth of CQF and SQF can also affect the performance. We search for the optimal value on the ICM→O experiment for the length of queries and the depth of Q-Former from two sets of changing values, such as $\left [8, 16, 32, 64 \right ]$ for the former, and $\left [1, 4, 8, 12 \right ]$ for the latter, respectively.

From Tab.~\ref{tab:Ablation_depth_length}, the following two conclusions can be drawn: (1) The length of the queries is set to around 16, achieving optimal performance. By observing at different depth settings, the number of learnable queries increases exponentially, the performance benefits of the model are subtle, and there is even a trend toward degradation. Specifically, at a length of 16, models at different depths achieve decent performance, with values of 3.33\%, 3.42\%, 3.56\%, and 3.33\% for HTER. (2) The depth of Q-Former has a negligible impact on the performance. In detail, under each length setting, the performance of models with different depths fluctuates around a certain value of HTER. For example, when the length is set to 8, the HTER is 3.45\%, while when the length is 16, 32, and 64, the HTER is 3.33\%. Based on the experimental results, we suggest setting the number of queries to 16 and the depth of the Q-Former to 1.

\subsection{Visualization and Analysis}
\begin{figure}[t]
	\centering
	\includegraphics[width=0.9\linewidth]{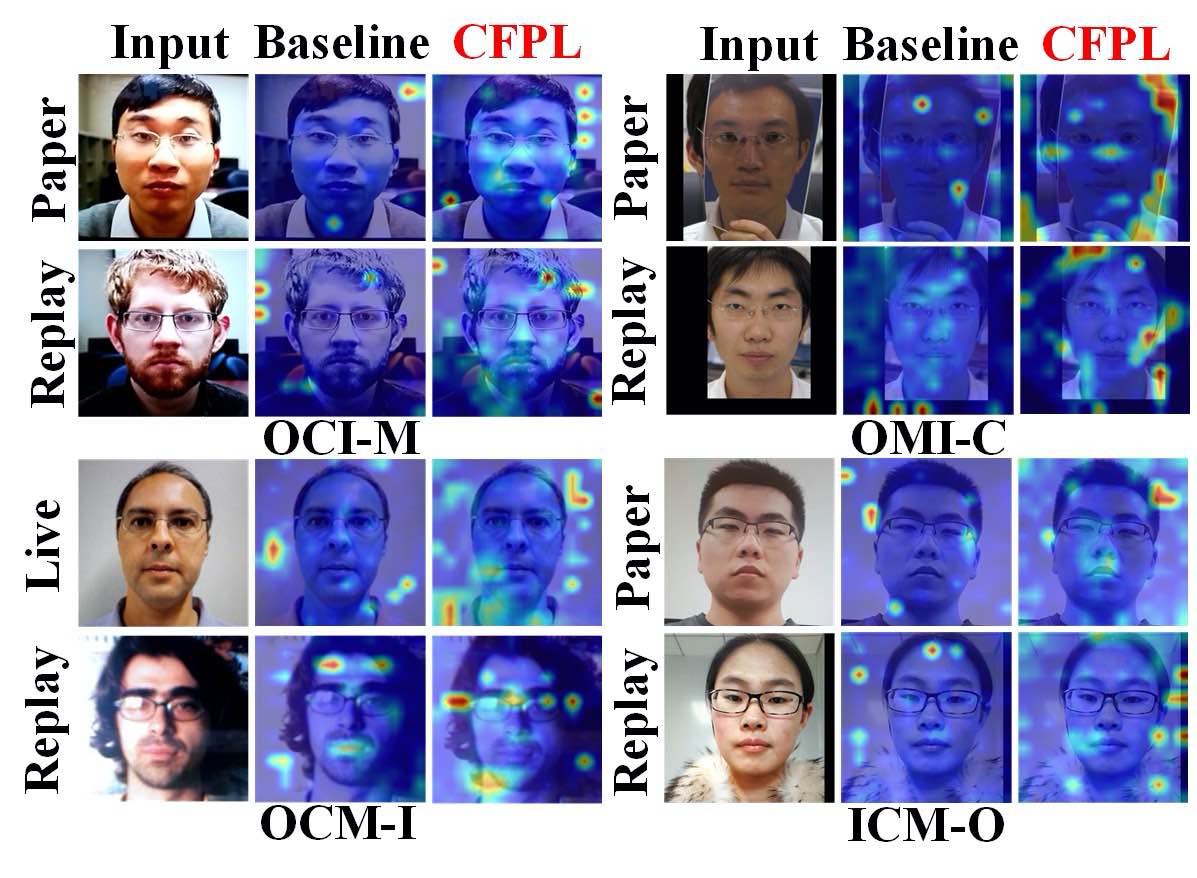}
	\vspace{-0.4cm}
	\caption{Using visualization tool~\cite{Chefer_2021_ICCV}, the attention maps on all sub-protocols from Protocol 1, where the Baseline caused classification errors due to its failure to detect spoofing regions, and our CFPL correctly classifies these samples by correcting the region of interest.
	}
	\label{fig:cfpl_map}
\end{figure}

With attention-model explainability tool~\cite{Chefer_2021_ICCV}, we visually validate the superiority of the proposed CFPL from the visual attention maps, compared to the Baseline, which is obtained by removing all contributions from the CFPL. The results on all protocols are shown in Fig.~\ref{fig:cfpl_map}, where the maps of the Baseline correspond to misclassified samples, while our CFPL correctly classifies these samples.

Specifically, for the OCM→I experiment on Protocol 1, the Baseline classifies live face errors due to focusing more on the background. Our CFPL correctly classifies it by correcting the focus area to the boundary between face and background. For the playback attack, the Baseline does not pay attention to spoofing clues, such as the reflection spot on the electronic screen, and the feature map area illuminated by our method. Similar conclusions can be drawn on other sub-protocols. 

\section{Conclusion}

In this work, we target DG FAS via textual prompt learning for the first time, and present a cross-domain FAS framework CFPL, which utilizes two lightweight transformers, CQF and SQF, to learn the different semantic prompts conditioned on content and style features. Finally, we introduce text supervision, diverse style prompt, and prompt modulation to promote the generalization. 


\section{Acknowledgments}
This work was supported by the National Key Research and Development Program of China under Grant 2021YFE0205700, Beijing Natural Science Foundation JQ23016, the Science and Technology Development Fund of Macau Project 0123/2022/A3, 0096/2023/RIA2, and 0070/2020/AMJ, Guangdong Provincial Key R\&D Programme: 2019B010148001, the Chinese National Natural Science Foundation Project 62276254, U23B2054, and the InnoHK program.

{
    \small
    \bibliographystyle{ieeenat_fullname}
    \bibliography{main}

\begin{thebibliography}{63}
\providecommand{\natexlab}[1]{#1}
\providecommand{\url}[1]{\texttt{#1}}
\expandafter\ifx\csname urlstyle\endcsname\relax
  \providecommand{\doi}[1]{doi: #1}\else
  \providecommand{\doi}{doi: \begingroup \urlstyle{rm}\Url}\fi

\bibitem[Boulkenafet et~al.(2017)Boulkenafet, Komulainen, Li, Feng, and
  Hadid]{Boulkenafet2017OULU}
Zinelabinde Boulkenafet, Jukka Komulainen, Lei Li, Xiaoyi Feng, and Abdenour
  Hadid.
\newblock Oulu-npu: A mobile face presentation attack database with real-world
  variations.
\newblock In \emph{FGR}, pages 612--618, 2017.

\bibitem[Chefer et~al.(2021)Chefer, Gur, and Wolf]{Chefer_2021_ICCV}
Hila Chefer, Shir Gur, and Lior Wolf.
\newblock Generic attention-model explainability for interpreting bi-modal and
  encoder-decoder transformers.
\newblock In \emph{Proceedings of the IEEE/CVF International Conference on
  Computer Vision (ICCV)}, pages 397--406, 2021.

\bibitem[Chen et~al.(2021)Chen, Yao, Sheng, Ding, Tai, Li, Huang, and
  Jin]{chen2021generalizable}
Zhihong Chen, Taiping Yao, Kekai Sheng, Shouhong Ding, Ying Tai, Jilin Li,
  Feiyue Huang, and Xinyu Jin.
\newblock Generalizable representation learning for mixture domain face
  anti-spoofing.
\newblock In \emph{Proceedings of the AAAI Conference on Artificial
  Intelligence}, pages 1132--1139, 2021.

\bibitem[Chingovska et~al.(2012)Chingovska, Anjos, and
  Marcel]{Chingovska_BIOSIG-2012}
Ivana Chingovska, Andr{\'e} Anjos, and S{\'e}bastien Marcel.
\newblock On the effectiveness of local binary patterns in face anti-spoofing.
\newblock In \emph{BIOSIG}, 2012.

\bibitem[Fang et~al.(2023{\natexlab{a}})Fang, Liu, Wan, Escalera, Escalante,
  and Lei]{Fang_2023_CVPR}
Hao Fang, Ajian Liu, Jun Wan, Sergio Escalera, Hugo~Jair Escalante, and Zhen
  Lei.
\newblock Surveillance face presentation attack detection challenge.
\newblock In \emph{Proceedings of the IEEE/CVF Conference on Computer Vision
  and Pattern Recognition (CVPR) Workshops}, pages 6360--6370,
  2023{\natexlab{a}}.

\bibitem[Fang et~al.(2023{\natexlab{b}})Fang, Liu, Wan, Escalera, Zhao, Zhang,
  Li, and Lei]{fang2023surveillance}
Hao Fang, Ajian Liu, Jun Wan, Sergio Escalera, Chenxu Zhao, Xu Zhang, Stan~Z
  Li, and Zhen Lei.
\newblock Surveillance face anti-spoofing.
\newblock \emph{IEEE Transactions on Information Forensics and Security},
  2023{\natexlab{b}}.

\bibitem[George and Marcel(2019)]{george2019deep}
Anjith George and S{\'e}bastien Marcel.
\newblock Deep pixel-wise binary supervision for face presentation attack
  detection.
\newblock In \emph{ICB}, 2019.

\bibitem[George and Marcel(2021)]{george2021cross}
Anjith George and S{\'e}bastien Marcel.
\newblock Cross modal focal loss for rgbd face anti-spoofing.
\newblock In \emph{CVPR}, pages 7882--7891, 2021.

\bibitem[George et~al.(2019)George, Mostaani, Geissenbuhler, Nikisins, Anjos,
  and Marcel]{george2019biometric}
Anjith George, Zohreh Mostaani, David Geissenbuhler, Olegs Nikisins, Andr{\'e}
  Anjos, and S{\'e}bastien Marcel.
\newblock Biometric face presentation attack detection with multi-channel
  convolutional neural network.
\newblock \emph{TIFS}, 2019.

\bibitem[Huang et~al.(2022)Huang, Sun, Liu, Chu, Xiao, Yuan, Adam, and
  Yang]{huang2022adaptive}
Hsin-Ping Huang, Deqing Sun, Yaojie Liu, Wen-Sheng Chu, Taihong Xiao, Jinwei
  Yuan, Hartwig Adam, and Ming-Hsuan Yang.
\newblock Adaptive transformers for robust few-shot cross-domain face
  anti-spoofing.
\newblock \emph{arXiv preprint arXiv:2203.12175}, 2022.

\bibitem[Huang and Belongie(2017)]{huang2017arbitrary}
Xun Huang and Serge Belongie.
\newblock Arbitrary style transfer in real-time with adaptive instance
  normalization.
\newblock In \emph{Proceedings of the IEEE international conference on computer
  vision}, pages 1501--1510, 2017.

\bibitem[Jia et~al.(2020)Jia, Zhang, Shan, and Chen]{jia2020single}
Yunpei Jia, Jie Zhang, Shiguang Shan, and Xilin Chen.
\newblock Single-side domain generalization for face anti-spoofing.
\newblock In \emph{Proceedings of the IEEE/CVF Conference on Computer Vision
  and Pattern Recognition}, pages 8484--8493, 2020.

\bibitem[Li et~al.(2021)Li, Selvaraju, Gotmare, Joty, Xiong, and
  Hoi]{li2021align}
Junnan Li, Ramprasaath Selvaraju, Akhilesh Gotmare, Shafiq Joty, Caiming Xiong,
  and Steven Chu~Hong Hoi.
\newblock Align before fuse: Vision and language representation learning with
  momentum distillation.
\newblock \emph{Advances in neural information processing systems},
  34:\penalty0 9694--9705, 2021.

\bibitem[Li et~al.(2023)Li, Li, Savarese, and Hoi]{li2023blip}
Junnan Li, Dongxu Li, Silvio Savarese, and Steven Hoi.
\newblock Blip-2: Bootstrapping language-image pre-training with frozen image
  encoders and large language models.
\newblock \emph{arXiv preprint arXiv:2301.12597}, 2023.

\bibitem[Liu and Liang(2022)]{ijcai2022p165}
Ajian Liu and Yanyan Liang.
\newblock Ma-vit: Modality-agnostic vision transformers for face anti-spoofing.
\newblock In \emph{Proceedings of the Thirty-First International Joint
  Conference on Artificial Intelligence, {IJCAI-22}}, pages 1180--1186.
  International Joint Conferences on Artificial Intelligence Organization,
  2022.

\bibitem[Liu et~al.(2021{\natexlab{a}})Liu, Tan, Wan, Escalera, Guo, and
  Li]{liu2021casia}
Ajian Liu, Zichang Tan, Jun Wan, Sergio Escalera, Guodong Guo, and Stan~Z Li.
\newblock Casia-surf cefa: A benchmark for multi-modal cross-ethnicity face
  anti-spoofing.
\newblock In \emph{Proceedings of the IEEE/CVF Winter Conference on
  Applications of Computer Vision}, pages 1179--1187, 2021{\natexlab{a}}.

\bibitem[Liu et~al.(2021{\natexlab{b}})Liu, Tan, Wan, Liang, Lei, Guo, and
  Li]{liu2021face}
Ajian Liu, Zichang Tan, Jun Wan, Yanyan Liang, Zhen Lei, Guodong Guo, and
  Stan~Z Li.
\newblock Face anti-spoofing via adversarial cross-modality translation.
\newblock \emph{IEEE Transactions on Information Forensics and Security},
  16:\penalty0 2759--2772, 2021{\natexlab{b}}.

\bibitem[Liu et~al.(2021{\natexlab{c}})Liu, Zhao, Yu, Su, Liu, Kong, Wan,
  Escalera, Escalante, Lei, et~al.]{liu20213d}
Ajian Liu, Chenxu Zhao, Zitong Yu, Anyang Su, Xing Liu, Zijian Kong, Jun Wan,
  Sergio Escalera, Hugo~Jair Escalante, Zhen Lei, et~al.
\newblock 3d high-fidelity mask face presentation attack detection challenge.
\newblock In \emph{Proceedings of the IEEE/CVF International Conference on
  Computer Vision Workshops}, pages 814--823, 2021{\natexlab{c}}.

\bibitem[Liu et~al.(2022{\natexlab{a}})Liu, Wan, Jiang, Wang, and
  Liang]{liu2022disentangling}
Ajian Liu, Jun Wan, Ning Jiang, Hongbin Wang, and Yanyan Liang.
\newblock Disentangling facial pose and appearance information for face
  anti-spoofing.
\newblock In \emph{2022 26th International Conference on Pattern Recognition
  (ICPR)}, pages 4537--4543. IEEE, 2022{\natexlab{a}}.

\bibitem[Liu et~al.(2022{\natexlab{b}})Liu, Zhao, Yu, Wan, Su, Liu, Tan,
  Escalera, Xing, Liang, et~al.]{liu2022contrastive}
Ajian Liu, Chenxu Zhao, Zitong Yu, Jun Wan, Anyang Su, Xing Liu, Zichang Tan,
  Sergio Escalera, Junliang Xing, Yanyan Liang, et~al.
\newblock Contrastive context-aware learning for 3d high-fidelity mask face
  presentation attack detection.
\newblock \emph{IEEE Transactions on Information Forensics and Security},
  17:\penalty0 2497--2507, 2022{\natexlab{b}}.

\bibitem[Liu et~al.(2023{\natexlab{a}})Liu, Tan, Liang, and Wan]{Liu_2023_CVPR}
Ajian Liu, Zichang Tan, Yanyan Liang, and Jun Wan.
\newblock Attack-agnostic deep face anti-spoofing.
\newblock In \emph{Proceedings of the IEEE/CVF Conference on Computer Vision
  and Pattern Recognition (CVPR) Workshops}, pages 6335--6344,
  2023{\natexlab{a}}.

\bibitem[Liu et~al.(2023{\natexlab{b}})Liu, Tan, Yu, Zhao, Wan, Liang, Lei,
  Zhang, Li, and Guo]{Liu2023FMViTFM}
Ajian Liu, Zichang Tan, Zitong Yu, Chenxu Zhao, Jun Wan, Yanyan Liang, Zhen
  Lei, Du Zhang, S. Li, and Guodong Guo.
\newblock Fm-vit: Flexible modal vision transformers for face anti-spoofing.
\newblock \emph{IEEE Transactions on Information Forensics and Security},
  18:\penalty0 4775--4786, 2023{\natexlab{b}}.

\bibitem[Liu et~al.(2023{\natexlab{c}})Liu, Tan, Chen, Wei, Zhao, and
  Wang]{liu2023unified}
Huan Liu, Zichang Tan, Qiang Chen, Yunchao Wei, Yao Zhao, and Jingdong Wang.
\newblock Unified frequency-assisted transformer framework for detecting and
  grounding multi-modal manipulation.
\newblock \emph{arXiv preprint arXiv:2309.09667}, 2023{\natexlab{c}}.

\bibitem[Liu et~al.(2023{\natexlab{d}})Liu, Tan, Tan, Wei, Zhao, and
  Wang]{liu2023forgery}
Huan Liu, Zichang Tan, Chuangchuang Tan, Yunchao Wei, Yao Zhao, and Jingdong
  Wang.
\newblock Forgery-aware adaptive transformer for generalizable synthetic image
  detection.
\newblock \emph{arXiv preprint arXiv:2312.16649}, 2023{\natexlab{d}}.

\bibitem[Liu et~al.(2024{\natexlab{a}})Liu, Liu, Tan, Li, and
  Zhao]{liu2024padvg}
Huan Liu, Xiaolong Liu, Zichang Tan, Xiaolong Li, and Yao Zhao.
\newblock Padvg: A simple baseline of active protection for audio-driven video
  generation.
\newblock \emph{ACM Transactions on Multimedia Computing, Communications and
  Applications}, 20\penalty0 (6), 2024{\natexlab{a}}.

\bibitem[Liu et~al.(2016)Liu, Yang, Yuen, and Zhao]{liu20163d2}
Siqi Liu, Baoyao Yang, Pong~C Yuen, and Guoying Zhao.
\newblock A 3d mask face anti-spoofing database with real world variations.
\newblock In \emph{CVPRW}, 2016.

\bibitem[Liu et~al.(2021{\natexlab{d}})Liu, Zhang, Yao, Bi, Ding, Li, Huang,
  and Ma]{liu2021adaptive}
Shubao Liu, Ke-Yue Zhang, Taiping Yao, Mingwei Bi, Shouhong Ding, Jilin Li,
  Feiyue Huang, and Lizhuang Ma.
\newblock Adaptive normalized representation learning for generalizable face
  anti-spoofing.
\newblock In \emph{Proceedings of the 29th ACM International Conference on
  Multimedia}, pages 1469--1477, 2021{\natexlab{d}}.

\bibitem[Liu et~al.(2021{\natexlab{e}})Liu, Zhang, Yao, Sheng, Ding, Tai, Li,
  Xie, and Ma]{liu2021dual}
Shubao Liu, Ke-Yue Zhang, Taiping Yao, Kekai Sheng, Shouhong Ding, Ying Tai,
  Jilin Li, Yuan Xie, and Lizhuang Ma.
\newblock Dual reweighting domain generalization for face presentation attack
  detection.
\newblock \emph{arXiv preprint arXiv:2106.16128}, 2021{\natexlab{e}}.

\bibitem[Liu and Liu(2022)]{liu2022spoof}
Yaojie Liu and Xiaoming Liu.
\newblock Spoof trace disentanglement for generic face anti-spoofing.
\newblock \emph{IEEE Transactions on Pattern Analysis and Machine
  Intelligence}, 45\penalty0 (3):\penalty0 3813--3830, 2022.

\bibitem[Liu et~al.(2018)Liu, Jourabloo, and Liu]{Liu2018Learning}
Yaojie Liu, Amin Jourabloo, and Xiaoming Liu.
\newblock Learning deep models for face anti-spoofing: Binary or auxiliary
  supervision.
\newblock In \emph{CVPR}, 2018.

\bibitem[Liu et~al.(2019)Liu, Stehouwer, Jourabloo, and Liu]{liu2019deep}
Yaojie Liu, Joel Stehouwer, Amin Jourabloo, and Xiaoming Liu.
\newblock Deep tree learning for zero-shot face anti-spoofing.
\newblock In \emph{CVPR}, 2019.

\bibitem[Liu et~al.(2020)Liu, Stehouwer, and Liu]{liu2020disentangling}
Yaojie Liu, Joel Stehouwer, and Xiaoming Liu.
\newblock On disentangling spoof trace for generic face anti-spoofing.
\newblock In \emph{ECCV}, pages 406--422. Springer, 2020.

\bibitem[Liu et~al.(2022{\natexlab{c}})Liu, Chen, Dai, Gou, Huang, and
  Xiong]{liu2022source}
Yuchen Liu, Yabo Chen, Wenrui Dai, Mengran Gou, Chun-Ting Huang, and Hongkai
  Xiong.
\newblock Source-free domain adaptation with contrastive domain alignment and
  self-supervised exploration for face anti-spoofing.
\newblock In \emph{ECCV}, 2022{\natexlab{c}}.

\bibitem[Liu et~al.(2022{\natexlab{d}})Liu, Chen, Dai, Li, Zou, and
  Xiong]{liu2022causal}
Yuchen Liu, Yabo Chen, Wenrui Dai, Chenglin Li, Junni Zou, and Hongkai Xiong.
\newblock Causal intervention for generalizable face anti-spoofing.
\newblock In \emph{ICME}, 2022{\natexlab{d}}.

\bibitem[Liu et~al.(2023{\natexlab{e}})Liu, Chen, Gou, Huang, Wang, Dai, and
  Xiong]{liu2023towards}
Yuchen Liu, Yabo Chen, Mengran Gou, Chun-Ting Huang, Yaoming Wang, Wenrui Dai,
  and Hongkai Xiong.
\newblock Towards unsupervised domain generalization for face anti-spoofing.
\newblock In \emph{Proceedings of the IEEE/CVF International Conference on
  Computer Vision}, 2023{\natexlab{e}}.

\bibitem[Liu et~al.(2024{\natexlab{b}})Liu, Chen, Dai, Gou, Huang, and
  Xiong]{liu2024source}
Yuchen Liu, Yabo Chen, Wenrui Dai, Mengran Gou, Chun-Ting Huang, and Hongkai
  Xiong.
\newblock Source-free domain adaptation with domain generalized pretraining for
  face anti-spoofing.
\newblock \emph{IEEE Transactions on Pattern Analysis and Machine
  Intelligence}, 2024{\natexlab{b}}.

\bibitem[Mostaani et~al.(2020)Mostaani, George, Heusch, Geissbuhler, and
  Marcel]{mostaani2020highquality}
Zohreh Mostaani, Anjith George, Guillaume Heusch, David Geissbuhler, and
  Sebastien Marcel.
\newblock The high-quality wide multi-channel attack (hq-wmca) database, 2020.

\bibitem[Qin et~al.(2021)Qin, Yu, Yan, Wang, Zhao, and Lei]{qin2021meta}
Yunxiao Qin, Zitong Yu, Longbin Yan, Zezheng Wang, Chenxu Zhao, and Zhen Lei.
\newblock Meta-teacher for face anti-spoofing.
\newblock \emph{IEEE transactions on pattern analysis and machine
  intelligence}, 2021.

\bibitem[Radford et~al.(2021)Radford, Kim, Hallacy, Ramesh, Goh, Agarwal,
  Sastry, Askell, Mishkin, Clark, et~al.]{radford2021learning}
Alec Radford, Jong~Wook Kim, Chris Hallacy, Aditya Ramesh, Gabriel Goh,
  Sandhini Agarwal, Girish Sastry, Amanda Askell, Pamela Mishkin, Jack Clark,
  et~al.
\newblock Learning transferable visual models from natural language
  supervision.
\newblock In \emph{International conference on machine learning}, pages
  8748--8763. PMLR, 2021.

\bibitem[Shao et~al.(2019)Shao, Lan, Li, and Yuen]{Rui2019Multi}
Rui Shao, Xiangyuan Lan, Jiawei Li, and Pong~C. Yuen.
\newblock Multi-adversarial discriminative deep domain generalization for face
  presentation attack detection.
\newblock In \emph{CVPR}, 2019.

\bibitem[Shao et~al.(2020)Shao, Lan, and Yuen]{shao2020regularized}
Rui Shao, Xiangyuan Lan, and Pong~C Yuen.
\newblock Regularized fine-grained meta face anti-spoofing.
\newblock In \emph{Proceedings of the AAAI Conference on Artificial
  Intelligence}, pages 11974--11981, 2020.

\bibitem[Srivatsan et~al.(2023)Srivatsan, Naseer, and
  Nandakumar]{srivatsan2023flip}
Koushik Srivatsan, Muzammal Naseer, and Karthik Nandakumar.
\newblock Flip: Cross-domain face anti-spoofing with language guidance.
\newblock In \emph{Proceedings of the IEEE/CVF International Conference on
  Computer Vision}, pages 19685--19696, 2023.

\bibitem[Sun et~al.(2023)Sun, Liu, Liu, Li, and Chu]{sun2023rethinking}
Yiyou Sun, Yaojie Liu, Xiaoming Liu, Yixuan Li, and Wen-Sheng Chu.
\newblock Rethinking domain generalization for face anti-spoofing: Separability
  and alignment.
\newblock In \emph{Proceedings of the IEEE/CVF Conference on Computer Vision
  and Pattern Recognition}, pages 24563--24574, 2023.

\bibitem[Tan et~al.(2023)Tan, Li, Zhou, Wan, Lei, and Zhang]{tan2023compound}
Hao Tan, Jun Li, Yizhuang Zhou, Jun Wan, Zhen Lei, and Xiangyu Zhang.
\newblock Compound text-guided prompt tuning via image-adaptive cues.
\newblock \emph{arXiv preprint arXiv:2312.06401}, 2023.

\bibitem[Wang et~al.(2022{\natexlab{a}})Wang, Lu, Yang, and
  Lai]{Wang_2022_CVPR}
Chien-Yi Wang, Yu-Ding Lu, Shang-Ta Yang, and Shang-Hong Lai.
\newblock Patchnet: A simple face anti-spoofing framework via fine-grained
  patch recognition.
\newblock pages 20281--20290, 2022{\natexlab{a}}.

\bibitem[Wang et~al.(2019)Wang, Han, Shan, and Chen]{wang2019improving}
Guoqing Wang, Hu Han, Shiguang Shan, and Xilin Chen.
\newblock Improving cross-database face presentation attack detection via
  adversarial domain adaptation.
\newblock In \emph{2019 International Conference on Biometrics (ICB)}, pages
  1--8. IEEE, 2019.

\bibitem[Wang et~al.(2020{\natexlab{a}})Wang, Han, Shan, and Chen]{Wangarticle}
Guoqing Wang, Hu Han, Shiguang Shan, and Xilin Chen.
\newblock Unsupervised adversarial domain adaptation for cross-domain face
  presentation attack detection.
\newblock \emph{TIFS}, 2020{\natexlab{a}}.

\bibitem[Wang et~al.(2021)Wang, Zhang, Bian, Cai, Wang, and Pu]{wang2021self}
Jingjing Wang, Jingyi Zhang, Ying Bian, Youyi Cai, Chunmao Wang, and Shiliang
  Pu.
\newblock Self-domain adaptation for face anti-spoofing.
\newblock In \emph{Proceedings of the AAAI Conference on Artificial
  Intelligence}, pages 2746--2754, 2021.

\bibitem[Wang et~al.(2020{\natexlab{b}})Wang, Yu, Zhao, Zhu, Qin, Zhou, Zhou,
  and Lei]{wang2020deep}
Zezheng Wang, Zitong Yu, Chenxu Zhao, Xiangyu Zhu, Yunxiao Qin, Qiusheng Zhou,
  Feng Zhou, and Zhen Lei.
\newblock Deep spatial gradient and temporal depth learning for face
  anti-spoofing.
\newblock In \emph{CVPR}, 2020{\natexlab{b}}.

\bibitem[Wang et~al.(2022{\natexlab{b}})Wang, Wang, Yu, Deng, Li, Gao, and
  Wang]{wang2022domain}
Zhuo Wang, Zezheng Wang, Zitong Yu, Weihong Deng, Jiahong Li, Tingting Gao, and
  Zhongyuan Wang.
\newblock Domain generalization via shuffled style assembly for face
  anti-spoofing.
\newblock In \emph{Proceedings of the IEEE/CVF Conference on Computer Vision
  and Pattern Recognition}, pages 4123--4133, 2022{\natexlab{b}}.

\bibitem[Wen et~al.(2015)Wen, Han, and Jain]{wen2015face}
Di Wen, Hu Han, and Anil~K Jain.
\newblock Face spoof detection with image distortion analysis.
\newblock \emph{IEEE TIFS}, 2015.

\bibitem[Yu et~al.(2020{\natexlab{a}})Yu, Wan, Qin, Li, Li, and
  Zhao]{yu2020nasfas}
Zitong Yu, Jun Wan, Yunxiao Qin, Xiaobai Li, Stan~Z. Li, and Guoying Zhao.
\newblock Nas-fas: Static-dynamic central difference network search for face
  anti-spoofing.
\newblock In \emph{TPAMI}, 2020{\natexlab{a}}.

\bibitem[Yu et~al.(2020{\natexlab{b}})Yu, Zhao, Wang, Qin, Su, Li, Zhou, and
  Zhao]{yu2020searching}
Zitong Yu, Chenxu Zhao, Zezheng Wang, Yunxiao Qin, Zhuo Su, Xiaobai Li, Feng
  Zhou, and Guoying Zhao.
\newblock Searching central difference convolutional networks for face
  anti-spoofing.
\newblock In \emph{CVPR}, 2020{\natexlab{b}}.

\bibitem[Yu et~al.(2023)Yu, Liu, Zhao, Cheng, Cheng, and
  Zhao]{yu2023flexiblemodal}
Zitong Yu, Ajian Liu, Chenxu Zhao, Kevin H.~M. Cheng, Xu Cheng, and Guoying
  Zhao.
\newblock Flexible-modal face anti-spoofing: A benchmark, 2023.

\bibitem[Zhang et~al.(2020{\natexlab{a}})Zhang, Yao, Zhang, Tai, Ding, Li,
  Huang, Song, and Ma]{zhang2020face}
Ke-Yue Zhang, Taiping Yao, Jian Zhang, Ying Tai, Shouhong Ding, Jilin Li,
  Feiyue Huang, Haichuan Song, and Lizhuang Ma.
\newblock Face anti-spoofing via disentangled representation learning.
\newblock In \emph{ECCV}, 2020{\natexlab{a}}.

\bibitem[Zhang et~al.(2020{\natexlab{b}})Zhang, Liu, Wan, Liang, Guo, Escalera,
  Escalante, and Li]{zhang2020casia}
Shifeng Zhang, Ajian Liu, Jun Wan, Yanyan Liang, Guodong Guo, Sergio Escalera,
  Hugo~Jair Escalante, and Stan~Z Li.
\newblock Casia-surf: A large-scale multi-modal benchmark for face
  anti-spoofing.
\newblock \emph{TBMIO}, 2\penalty0 (2):\penalty0 182--193, 2020{\natexlab{b}}.

\bibitem[Zhang et~al.(2020{\natexlab{c}})Zhang, Yin, Li, Yin, Yan, Shao, and
  Liu]{CelebA-Spoof}
Yuanhan Zhang, Zhenfei Yin, Yidong Li, Guojun Yin, Junjie Yan, Jing Shao, and
  Ziwei Liu.
\newblock Celeba-spoof: Large-scale face anti-spoofing dataset with rich
  annotations.
\newblock In \emph{ECCV}, 2020{\natexlab{c}}.

\bibitem[Zhang et~al.(2012)Zhang, Yan, Liu, Lei, Yi, and Li]{Zhang2012A}
Zhiwei Zhang, Junjie Yan, Sifei Liu, Zhen Lei, Dong Yi, and Stan~Z Li.
\newblock A face antispoofing database with diverse attacks.
\newblock In \emph{ICB}, 2012.

\bibitem[Zhou et~al.(2021)Zhou, Yang, Qiao, and Xiang]{zhou2021domain}
Kaiyang Zhou, Yongxin Yang, Yu Qiao, and Tao Xiang.
\newblock Domain generalization with mixstyle.
\newblock \emph{arXiv preprint arXiv:2104.02008}, 2021.

\bibitem[Zhou et~al.(2022{\natexlab{a}})Zhou, Yang, Loy, and
  Liu]{zhou2022cocoop}
Kaiyang Zhou, Jingkang Yang, Chen~Change Loy, and Ziwei Liu.
\newblock Conditional prompt learning for vision-language models.
\newblock In \emph{IEEE/CVF Conference on Computer Vision and Pattern
  Recognition (CVPR)}, 2022{\natexlab{a}}.

\bibitem[Zhou et~al.(2022{\natexlab{b}})Zhou, Yang, Loy, and Liu]{zhou2022coop}
Kaiyang Zhou, Jingkang Yang, Chen~Change Loy, and Ziwei Liu.
\newblock Learning to prompt for vision-language models.
\newblock \emph{International Journal of Computer Vision (IJCV)},
  2022{\natexlab{b}}.

\bibitem[Zhou et~al.(2022{\natexlab{c}})Zhou, Zhang, Yao, Yi, Ding, and
  Ma]{zhou2022adaptive}
Qianyu Zhou, Ke-Yue Zhang, Taiping Yao, Ran Yi, Shouhong Ding, and Lizhuang Ma.
\newblock Adaptive mixture of experts learning for generalizable face
  anti-spoofing.
\newblock In \emph{Proceedings of the 30th ACM International Conference on
  Multimedia}, pages 6009--6018, 2022{\natexlab{c}}.

\bibitem[Zhou et~al.(2023)Zhou, Zhang, Yao, Lu, Yi, Ding, and
  Ma]{zhou2023instance}
Qianyu Zhou, Ke-Yue Zhang, Taiping Yao, Xuequan Lu, Ran Yi, Shouhong Ding, and
  Lizhuang Ma.
\newblock Instance-aware domain generalization for face anti-spoofing.
\newblock In \emph{Proceedings of the IEEE/CVF Conference on Computer Vision
  and Pattern Recognition}, pages 20453--20463, 2023.

\end{thebibliography}
}


\end{document}